\documentclass[runningheads]{llncs}
\usepackage{graphicx}
\usepackage{amsmath,amssymb} 
\usepackage[width=122mm,left=12mm,paperwidth=146mm,height=193mm,top=12mm,paperheight=217mm]{geometry}

\usepackage{color,soul}
\usepackage[dvipsnames]{xcolor}
\usepackage{subfig}
\usepackage{animate}

\usepackage[pagebackref=true,breaklinks=true,letterpaper=true,colorlinks,bookmarks=false]{hyperref}

\begin{document}
\pagestyle{headings}
\mainmatter

\title{Viewpoint Estimation---Insights \& Model} 

\titlerunning{Viewpoint Estimation---Insights \& Model}

\authorrunning{Gilad Divon, Ayellet Tal}
\author{Gilad Divon, Ayellet Tal}
\institute{The Technion - Israel}
\maketitle

\begin{figure}[h]
\centering
\begin{tabular}{cc}
\includegraphics[width=0.48\textwidth,height=0.17\textheight]{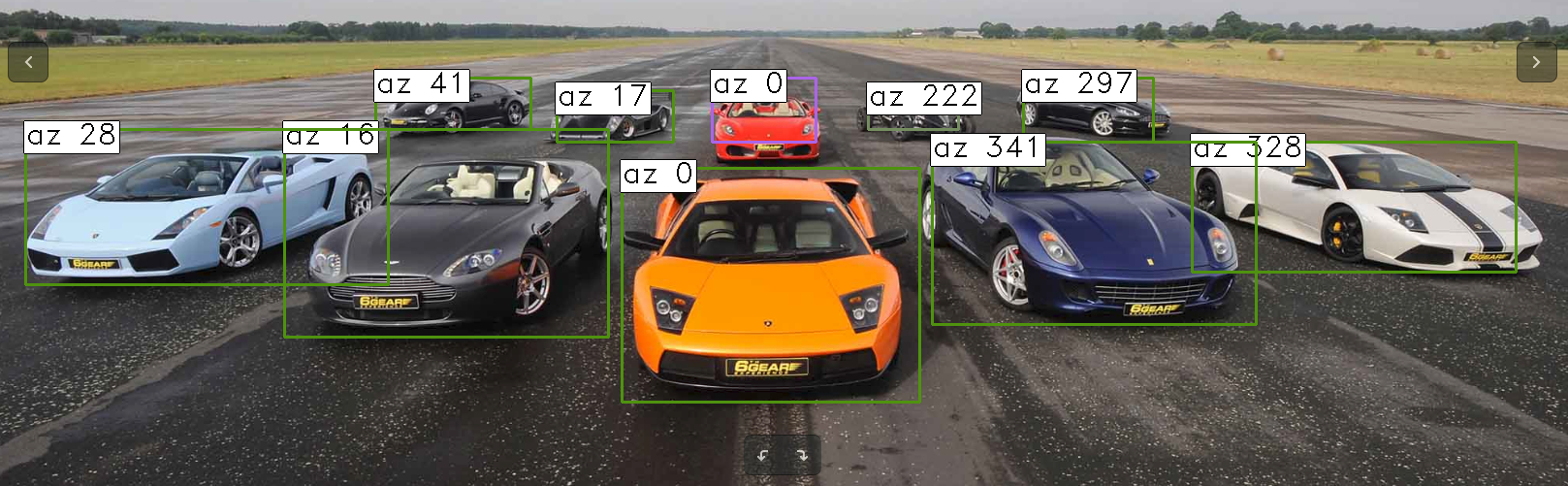}
&
\animategraphics[autoplay,loop,width=0.48\textwidth,keepaspectratio]{2}{./}{3}{31}\\
\end{tabular}
\vspace*{-0.3cm}
    \caption{\textbf{Viewpoint estimation.} 
    Given an image  containing objects from known categories, our model estimates the viewpoints (azimuth) of the objects.
    The image to the right is an animation that may be viewed in Adobe Acrobat Reader.}%
    \label{fig:teaser}%
\end{figure}

\vspace*{-1cm}
\begin{abstract}
This paper addresses the problem of viewpoint estimation of an object in a given image. 
It presents five key insights that should be taken into consideration when designing a CNN that solves the problem.
Based on these insights, the paper proposes a network in which
(i) The architecture jointly solves detection, classification, and viewpoint estimation.
(ii) New types of data are added and trained on.
(iii) A novel loss function, which takes into account both the geometry of the problem and the new types of data, is propose.
Our network improves the state-of-the-art results for this problem by 9.8\%.
\end{abstract}

\section{Introduction}
\label{sec:introduction}
Object category viewpoint estimation refers to the task of determining the viewpoints of objects in a given image, where the objects belong to known categories, as illustrated  in Figure.\ref{fig:teaser}.
This problem is an important component in our attempt to understand the 3D world around us and is therefore a long-term challenge in computer vision~\cite{huttenlocher1987old1,lowe1987old2,lowe1991old3,huttenlocher1990old4}, having numerous application~\cite{choi2012robot1,marchand2016_AR}.
The difficulty in solving the problem stems from the fact that a single image, which is a projection from 3D, does not yield sufficient information to determine the viewpoint.
Moreover, this problem suffers from scarcity of images with accurate viewpoint annotation, due not only to the high cost of manual annotation, but mostly to the imprecision of humans when estimating viewpoints.

Convolutional Neural Networks were recently applied to viewpoint estimation~\cite{su2015renderforCNN,massa2016crafting,tulsiani2015viewpoints}, leading to large improvements of state-of-the-art results on PASCAL3D+. 
Two major approaches were pursued.
The first is a regression approach, which handles the continuous values of viewpoints naturally~\cite{regression1,regression2,massa2016crafting}. 
This approach manages to represent the periodic characteristic of the viewpoint and is invertible.
However, as pointed out by~\cite{su2015renderforCNN}, the main limitation of regression for viewpoint estimation is that it cannot represent well the ambiguities that exist between different viewpoints of objects that have symmetries or near symmetries.

The second approach is to treat viewpoint estimation as a classification problem~\cite{su2015renderforCNN,tulsiani2015viewpoints}. 
In this case, viewpoints are transformed into a discrete space, where each viewpoint (angle) is represented as a single class (bin).
The network predicts the probability of an object to be in each of these classes.
This approach is shown to outperform regression, to be more robust, and to handle ambiguities better.
Nevertheless, its downside is that similar viewpoints are located in different bins and therefore, the bin order becomes insignificant.
This means that when the network errs, there is no advantage to small errors (nearby viewpoints) over large errors, as should be the case.
We follow the second approach.
We present five key insights, some of which were discussed before:
(i) Rather than separating the tasks of object detection, object classification, and viewpoint estimation, these should be integrated into a unified framework.
(ii)~As one of the major issues of this problem is the lack of labeled real images, novel ways to augment the data should be developed.
(iii)~The loss should reflect the geometry of the problem.
(iv)~Since viewpoints, unlike object classes, are related to one another, integrating over viewpoint predictions should outperform the selection of the strongest activation.
(v)~CNNs for viewpoint estimation improve as CNNs for object classification/detection do.

Based on these observations, we propose a network that improves the state-of-the-art results by 9.8\%, from $36.1\%$ to $45.9\%$, on PASCAL3D+~\cite{xiang_wacv14pascal3d}.
We touch each of the three components of any learning problem: architecture, data, and loss.
In particular, our architecture unifies object detection, object classification, and viewpoint estimation and is built on top of Faster R-CNN.
Furthermore, in addition to real and synthetic images, we also use flipped images and videos, in a semi-supervised manner.
This not only augments the data for training, but also lets us refine our loss.
Finally, we define a new loss function that reflects both the geometry of the problem and the new types of training data.

Therefore, this paper makes two major contributions.
First, it presents insights that should be the basis for viewpoint estimation algorithms (Section~\ref{sec:instights}).
Second, it presents a network (Section~\ref{sec:algorithm}) that achieves SOTA results (Section~\ref{section:results}).

\section{Our Insights in a nutshell}
\label{sec:instights}

We start our study with short descriptions of five insights we make on viewpoint estimation. 
In the next section, we introduce an algorithm that is based on these insights and generates state-of-the-art results.


\vspace{0.02in}
\noindent
1. \textit{Rather than separating the tasks of object detection, object classification, and viewpoint estimation, these should be integrated into a unified network.}
In~\cite{su2015renderforCNN}, an off-the-shelf R-CNN~\cite{girshick2014rcnn} was used.
Given the detection results, a network was designed to estimate the viewpoint.
In~\cite{massa2016crafting} classification and viewpoint estimation were solved jointly, while relying on bounding box suggestions from Deep Mask~\cite{pinheiro2015deepmask}/Fast R-CNN~\cite{girshick2015fast}.
%
We propose a different architecture that combines the three tasks and show that training the network jointly is beneficial.
This insight is in accordance with similar observations made in other domains~\cite{ren2015faster,eigen2015predicting,gkioxari2015contextual}.

\vspace{0.02in}
\noindent
2. {\it As one of the major issues of viewpoint estimation is the lack of labeled real images, novel ways to augment the data are necessary.}
In~\cite{su2015renderforCNN,massa2016crafting} it was proposed to use both real data and images of CAD models, for which backgrounds were randomly synthesized.
We propose to add two new types of training data, which not only increase the volume of data, but also benefit learning.
First, we horizontally flip the real images.
Since the orientation of these images is known, yet no new information regarding detection and classification is added, they are used within a new loss function to focus on viewpoint estimation.
Second,  we use unlabeled videos of objects for which, though we do not know the exact orientation, we do know that subsequent frames should be associated with nearby viewpoints.
This constraint is utilized to gain better viewpoint predictions.
Finally, as a minor modification, rather than randomly choosing backgrounds for the synthetic images, we choose backgrounds that suit the objects, e.g. backgrounds of the ocean should be added to boats, but not to airplanes.

\vspace{0.02in}
\noindent
3. {\it The loss should reflect the geometry of the problem, since viewpoint estimation is essentially a geometric problem, having geometric constraints.}
In~\cite{su2015renderforCNN}, the loss considers the geometry by giving larger weights to bins of close viewpoints.
In~\cite{massa2016crafting}, it was found that this was not really helpful and viewpoint estimation was solved purely as a classification problem.
We show that geometric constraints are very helpful.
Indeed, our loss function considers 
(1) the relations between the geometries of triplets of images,
(2) the constraints posed by the flipped images, 
and (3) the constraints posed by subsequent frames within videos

\vspace{0.02in}
\noindent
4. {\it 
An integration of the results is helpful.}
Previous work chose as the final result the bin that contains the viewpoint having the strongest activation.
Instead, we integrate over all the viewpoints within a bin and choose as the final result the bin that maximizes this integral.

\vspace{0.02in}
\noindent
5. \textit{As object classification/detection CNNs improve, so do CNNs for viewpoint estimation.}
%
In~\cite{su2015renderforCNN} AlexNet~\cite{krizhevsky2012imagenet} was used as the base network, whereas in~\cite{tulsiani2015viewpoints,massa2016crafting} VGG~\cite{simonyan2014VGG} was used.
Instead, we use ResNet~\cite{he2016resnet}.
This is not only because of its better performance in classification, but also due to its skip-connections concept.
These connections enable the flow of information between non-adjacent layers and by doing so, preserve spatial information from different scales. 
This idea is similar to the multi-scale approach of~\cite{tulsiani2015viewpoints}, which was shown benefit viewpoint estimation.


%
\vspace{0.02in}
\noindent
{\bf A concise view on the contribution of the insights:} Table~\ref{tbl:comparison_summary} summarizes the influence of each of our insights on the performance of viewpoint estimation. 
Our results are compared to the state-of-the-art results of~\cite{tulsiani2015viewpoints,massa2016crafting,su2015renderforCNN}.
The total gain of our algorithm is $9.8\%$ compared to~\cite{massa2016crafting}.
Section~\ref{section:results} will analyze these results in depth.  

\begin{table}[t]
\begin{center}
\begin{tabular}{|l|c|}
\hline
\hspace{0.9in}Method & Score (mAVP24)\\
\hline\hline
{\color{RoyalPurple}\cite{su2015renderforCNN}}: {\color{teal}AlexNet}-{\color{violet}geometry}-{\color{olive}synthetic+real}  & 19.8 \\
{\color{RoyalPurple}\cite{tulsiani2015viewpoints}}: {\color{teal}VGG}-{\color{violet}classification}-{\color{olive}real}  & 31.1 \\
{\color{RoyalPurple}\cite{massa2016crafting}}: {\color{teal}VGG}-{\color{violet}classification}-{\color{olive}synthetic+real} & 36.1 \\
\hline
{\color{RoyalPurple}Ours}: {\color{teal}Insights 1,5} - Architecture  & 40.6 \\
{\color{RoyalPurple}Ours}: {\color{teal}Insights: 1,4,5} - Integration & 43.2 \\
{\color{RoyalPurple}Ours}: {\color{teal}Insights: 1,3,4,5} - Loss  & 44.4 \\
{\color{RoyalPurple}Ours}: {\color{teal}Insights: 1,2,3,4,5} - Data &\textbf{45.9}\\
\hline
\end{tabular}
\end{center}
\vspace*{-0.3cm}
\caption{{\bf Contribution of the insights.} 
This table summarizes the influence of our insights on the performance. 
The total gain is $9.8\%$ compared to~\cite{massa2016crafting}.}
\label{tbl:comparison_summary}
\vspace*{-0.3cm}
\end{table}

\section{Model}
\label{sec:algorithm}
\begin{figure}[h]
\centering
\begin{tabular}{ccc}
 \includegraphics[width=0.3\textwidth,height=0.18\textheight]{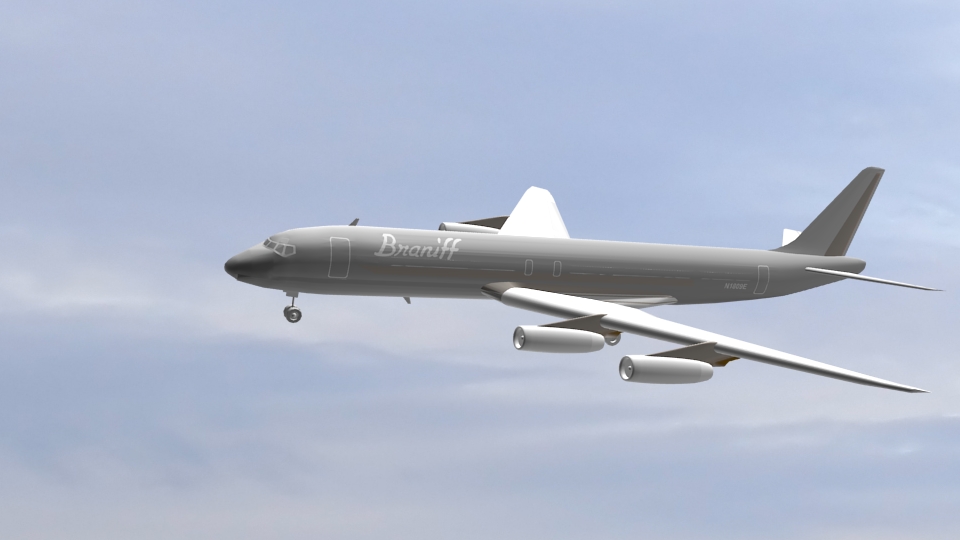} &
  \includegraphics[width=0.1\textwidth,height=0.18\textheight]{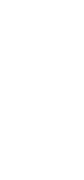} &
 \includegraphics[width=0.3\textwidth]{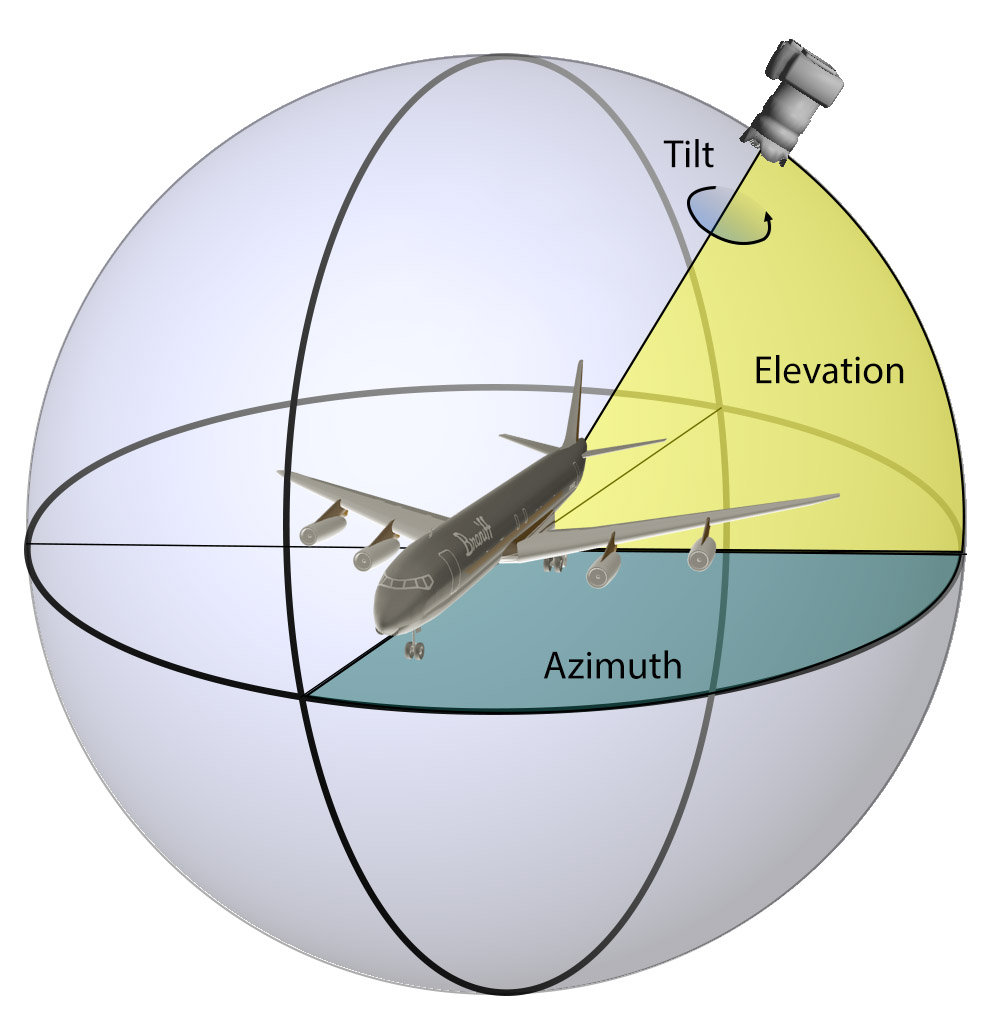}\\
 (a) Input && (b) Output
\end{tabular}
\vspace*{-0.2cm}
   \caption{\textbf{Problem definition}. 
   Given an image that contains an object (a), the goal is to retrieve the Euler angles that define the orientation of the camera relative to the object (b).}
    \label{fig:euler_angles}
    \vspace*{-0.3cm}
\end{figure}

Recall that we adopt the approach of treating viewpoint estimation as a classification problem.
Though a viewpoint is defined as a 3D vector, representing the camera orientation relative to the object (Figure~\ref{fig:euler_angles}), we focus on the azimuth, as  done in previous works; finding the other angles is equivalent.
The set of possible viewpoints is discretized into $360$ classes, where each class represents~${1^\circ}$.
This section presents the different components of our suggested network, which realizes the insights described in the previous section.

\subsection{Architecture}
\label{subsection: Architecture}

\begin{figure}[t]
\begin{center}
   \includegraphics[width=0.7\textwidth,height=0.3\textheight]{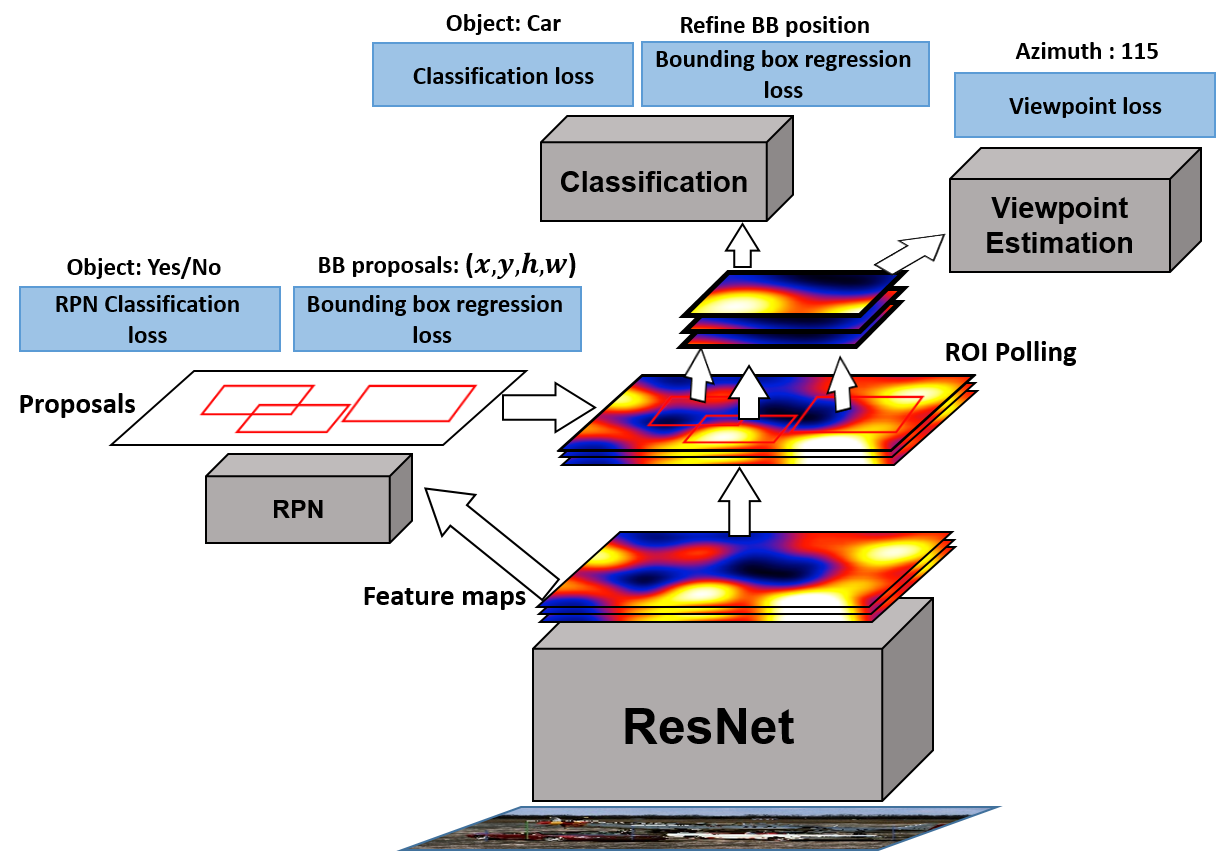}
\end{center}
\vspace*{-0.3cm}
   \caption{\textbf{Network architecture.}
   Deep features are extracted by ResNet and  passed to RPN to predict bounding boxes.
   After ROI pooling, they are passed both to the classification head and to the viewpoint estimation head. 
   The output consists of a set of bounding boxes $(x,y,h,w)$, and for each of them---the class of the object within the bounding box and its estimated viewpoint.}
\label{fig:architecture}
\end{figure}

Hereafter, we describe the implementation of Insights 1,4\&5, focusing on the integration of  classification, object detection and viewpoint estimation.
%
%
%
Figure~\ref{fig:architecture} sketches our general architecture.
It is based on Faster R-CNN~\cite{ren2015faster}, which both detects and classifies.
As a base network within Faster R-CNN, we use ResNet~\cite{he2016resnet}, which is shown to achieve better results for classification than VGG.
Another advantage of ResNet is its skip connections. 
To understand their importance, recall that in contrast to our goal, classification networks are trained to ignore viewpoints.
Skip connection allow the data to flow directly, without being distorted by 
pooling, which is known to disregard the inner order of activations.

A viewpoint estimation head is added on top of Faster R-CNN.
It is built similarly to the classification head, except for the size of the fully-connected layer, which is 4320 (the number of object classes * $360$ angles).
%

The resulting feature map of ResNet is passed to all the model's components: to the {\em Region Proposal Network (RPN)} of Faster R-CNN, which predict bounding boxes, to the classification component, and to the viewpoint estimation head.
The bounding box proposals are used to define the pooling regions that are input both to the classification head and to the viewpoint estimation head.
The latter outputs for each bounding box a vector, whose entries each represents a viewpoint prediction, assuming that the object in the bounding box belongs to a certain class, e.g. entries 0-359 are the predictions for boats, 360-719 for bicycles etc.
The relevant section of this vector is chosen as the output once the object class is predicted by the classification head.
The final output of the system is a set of bounding boxes $(x,y,h,w)$, and for each of them---the class of the object in the bounding box and object's viewpoint for this class---integrating the results of the classification head and the viewpoint estimation head.

\vspace{0.1in}
\noindent
{\bf Implementation details:}
Within this general framework, three issues should be addressed.
First, though viewpoint estimation is defined as a classification problem, we cannot simply use the classification head  of Faster R-CNN as is for the viewpoint estimation task.
This is so since the periodical pooling layers within the network are invariant to the location of the activation in the feature map. 
This is undesirable when evaluating an object's viewpoint, since different viewpoints have the same representation after pooling that uses Max or Average.
To solve this problem, while still accounting for the importance of the pooling layers, we replace only the last pooling layer of  the viewpoint estimation head with a fully connected layer (of size 1024).
This preserves the spatial information, as different weights are assigned to the different locations in the feature map. 

Second, in the original Faster R-CNN, the bounding box proposals are passed to a non-maximum suppression function in order to reduce the overlapping bounding box suggestions.
Bounding boxes whose {\em Interaction over Union (IoU)} is larger than $0.5$ are grouped together and the output is the bounding box with the highest prediction score.
Which viewpoint should be associated with this representative bounding box?

One option is to choose the angle of the selected bounding box ($BB$).
This, however, did not yield good results.
Instead, we compute the viewpoint vector (in which every possible viewpoint has a score) of $BB$ as follows.
Our network computes for each bounding box $bb_i$ a distribution of viewpoints $P_A(bb_i)$ and a classification score $P_C(bb_i)$.
We compute the distribution of the viewpoints for $BB$ by summing over the contributions of all the overlapping bounding boxes, weighted by their classification scores:
\begin{equation} 
viewpoint\_Score(BB) = \Sigma_{i}{P_A(bb_i)P_C(bb_i)}. 
\label{eq:viewpoint_score}
\end{equation}
This score vector, of length 360, is associated with $BB$.
Hence, our approach considers the predictions for all the bounding boxes when selecting the viewpoint.

Given this score vector, the viewpoint should be estimated.
The score is computed by summing Equation~\eqref{eq:viewpoint_score} over all the viewpoints within a bin.
Following~\cite{su2015renderforCNN,massa2016crafting}, this is done for $K=24$ bins, each representing $15^\circ$ angles,.
Then, the bin selected is the one for which this sum is maximized.
%

Third, we noticed that small objects are consistently mis-detected by Faster R-CNN, whereas such object do exist in our dataset.
To solve it, a minor modification was applied to the network.
We added a set of anchors of size $64$ pixels, in addition to the existing sizes of $\{128,256,512\}$ (anchors are the initial suggestions for bounding boxes sizes).
This led to a small increase of training time, but significantly improved the detection results (from 74.3\% to 77.8\% using mAP) and consequently improved the viewpoint estimation.

\subsection{Data}

In our problem, we need not only to classify objects, but also to sub-classify each object into viewpoints.
This means that a huge number of parameters must be learned, and this in turn requires a large amount of labeled data.
Yet, labeled real images are scarce, since viewpoint labeling is extremely difficult.

In~\cite{xiang_wacv14pascal3d}, a creative procedure was proposed:
Given a detected and classified object in an image, the user selects the most similar 3D CAD model (from Google 3D Warehouse~\cite{Google3DWarehouse}) and marks some corresponding key points. 
The 3D viewpoint is then computed for this object.
Since this procedure is expensive, the resulting dataset contains only 30K annotated images that belong to 12 categories.
This is the largest dataset with ground truth available today for this task.


To overcome the challenges of training data scarcity, Su {\em et al.}~\cite{su2015renderforCNN} proposed to augment the dataset with synthetic rendered CAD models from ShapeNet~\cite{shapenet2015}. 
This  allows the creation of as many images as needed for a single model. 
Backgrounds were added to the rendered  images by randomly selecting them from images of SUN397~\cite{xiao2010sun}.
The images were then cropped to resemble real images taken "in the wild", where the cropping statistics maintained that of VOC2012~\cite{everingham2010pascalVOC12}, creating 2M images.
The use of this synthetic data increased the performance by $\sim 2\%$.

We further augmented the training dataset, in accordance with Insight 2, in three manners.
First, rather than randomly selecting backgrounds, we chose for each category backgrounds that are realistic for the objects.
For instance, boats should not float in living-rooms, but rather be synthesized with backgrounds of oceans or harbors.
This change increased the performance only slightly.

More importantly, we augmented the training dataset by horizontally flipping the existing real images.
Since the orientation of these images is known, they are used within a new loss function to enforce correct viewpoints (Section~\ref{subsec:loss}).

Finally, we used unlabeled videos of objects, for which we could exploit the coherency of the motion, to further increase the volume of data and improve the results.
We will show in Section~\ref{subsec:loss} how to modify the loss function to use these clips for semi-supervised learning. 

\subsection{Loss}
\label{subsec:loss}
As shown in Figure~\ref{fig:architecture}, there are five loss functions in our model, four of which are set by Faster R-CNN.
This section focuses on the {\em viewpoint loss} function, in line of Insights 3 \& 4, and show how to combine it with the other loss functions.

Treating viewpoint estimation as a classification problem, the network predicts the probability of an object to belong to a viewpoint bin (bin$=1^\circ$).
One problem with this approach is that similar viewpoints are located in different bins and bin order is disregarded.
In the evaluation, however, the common practice is to divide the space of viewpoints into larger bins (of $15^\circ$)~\cite{xiang_wacv14pascal3d}.
This means that, in contrast to classical classification, if the network errs when estimating the viewpoint, it is better to err by outputting close viewpoints than by outputting faraway ones.
Therefore, our loss should address a geometric constraint---the network should produce similar representations for close viewpoints.

\begin{figure}[t]
\centering
   \includegraphics[height=4.5cm]{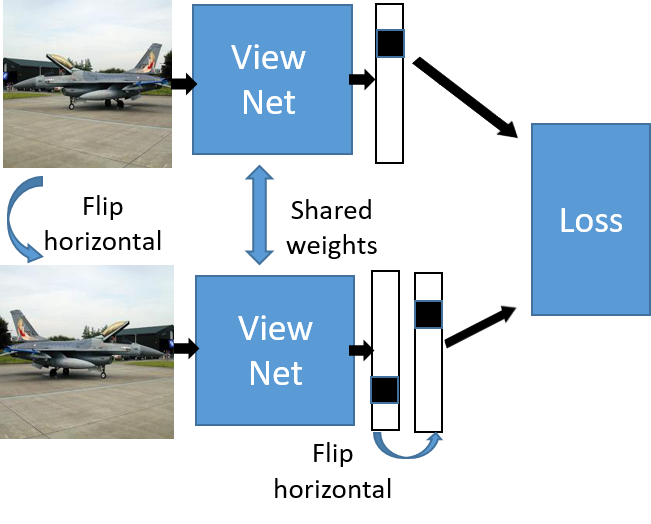}
\vspace*{-0.3cm}
   \caption{\textbf{Flipped images within a Siamese network}. 
   The loss attempts to minimize the distance between the representations of an image and the flip  of its flipped image.}
\label{fig:Siamese net}
\end{figure}

In address this, Su {\em et al.}~\cite{su2015renderforCNN} proposed to use a geometric-aware loss function instead of a regular cross entropy loss with one-hot label:
\begin{equation}
L_{geom}(\vec{q}) = -\frac{1}{C}\sum_{k=1}^{360} exp(-\frac{|k_{gt}-k|}{\sigma})log(q(k)).
\label{eq:geometricLoss}
\end{equation}
In this equation, $\vec{q}$ is the viewpoint probability vector of some bounding box,  $k$ is a bin index (and $k_{gt}$ is the ground truth index), $q(k)$ is the probability of bin $k$, and  $\sigma=3$.
Thus, in Equation~\eqref{eq:geometricLoss} the commonly-used one-hot label is replaced by an exponential decay weight w.r.t the distance between the viewpoints.
By doing so, correlation between predictions of nearby views is "encouraged".
Interestingly, while this loss function was shown to improve the results of~\cite{su2015renderforCNN}, it did not improve the results of a later work of~\cite{massa2016crafting}.

We propose a different loss function, which realizes the geometric constraint.
Our loss is based on the fundamental idea of the Siamese architecture~\cite{bromley1994signature,siamese2,siamese3}, which has the property of bringing similar classes closer together, while increasing the distances between unrelated classes.

Our first attempt was to utilize the contrastive Siamese loss~\cite{siamese2}, which is applied to the embedded representation of the viewpoint estimation head (before the viewpoint classification layer).
Given representations of two images $F(X_1),F(X_2)$ and the $L_2$ distance between them 
$D(X_1,X_2) = ||F(X_1)-F(X_2)||_2$, the loss is defined as:
\begin{equation}
L_{contrastive}(D) = (Y)\frac{1}{2}D^2 + (1-Y)\frac{1}{2}\left\{max(0,m-D)\right\}^2.
\label{eq:contrastive}
\end{equation}
Here, $Y$ is the similarity label, i.e. $1$ if the images have close viewpoints (in practice, up to $10^\circ$) and $0$ otherwise and $m$ is the margin.
Thus, pairs whose distance is larger than $m$  will not contribute to the loss.
There are two issues that should be addressed when adopting this loss: the choice of the hyper-parameter $m$ and the correct balance between the positive training examples and the negative ones, as this loss is sensitive to their number and to their order.
This approach yielded sub-optimal results for a variety of choices of $m$ and numbers/orders.

Therefore, we propose a different \& novel Siamese loss, as illustrated in Figure~\ref{fig:Siamese net}.
The key idea is to use pairs of an image and its horizontally-flipped image.
Since the only difference between these images is the viewpoint and the relation between the viewpoints is known, we define the following loss function:
\begin{equation}
  L_{flip}(X,X_{flip}) = L_{geom}(X)+L_{geom}(X_{flip}) +\lambda ||F(X)-flip(F(X_{flip}))||^2_2,
      \label{eq:Lflip}
\end{equation}
where $L_{geom}$ is from Equation~\eqref{eq:geometricLoss}.
 We expect the $L_2$ distance term, between the embeddings of an image and the flip of its flipped image, to be close to $0$.

To improve the results further, we adopt the triplet network concept~\cite{triplet,triplet2} and modify its loss to suit our problem. 
The basic idea is to "encourage" the network to output similarity-induced embeddings.
Three images are provided during training: $X^{ref},X^+,X^-$, where $X^{ref},X^+$ are from similar classes and $X^{ref},X^-$ are from dissimilar classes.
In~\cite{triplet}, the distance between image representations $D(F(X_1),F(X_2))$ is the $L_2$ distance between them.
Let $D^+= D(X^{ref},X^+)$, $D^-= D(X^{ref},X^-)$, and  $d^+,d^-$  be the results of applying softmax to $D^+, D^-$ respectively.
The larger the difference between the viewpoints, the more dissimilar the classes should be, i.e. $D^+ < D^-$.

A common loss, which encourages embeddings of related classes to have small distances and embeddings of unrelated classes to have large distances, is:
\begin{equation}
L_{triplet}(X^{ref},X^+,X^+) = ||(d^+,1-d^-)||_2^2.
\label{eq:Ltriplet}
\end{equation}

We found, however, that the distances $D$ get very large values and therefore, applying softmax to them results in $d^+,d^-$ that are very far from each other, even for similar labels.
Therefore, we replace $D$ by the {\em cosine} distance:
\begin{equation}
D(F(x_1),F(x_2)) = \frac{F(x_1)\cdot F(x_2)}{||F(x_1)||_2||F(x_2)||_2}.
\label{eq:cosine_dist}
\end{equation}
The distances are now in the range $[-1,1]$, which allows faster training and convergence, since the network does not need to account for changes in the scale of the weights.
For {\em cosine} distance we require  $D^+ > D^-$ (instead of $<$), and consequentially the roles of $d^+,d^-$ in Equation~\eqref{eq:Ltriplet} should switch.

A minor trick we apply for softmax to produce the range $[0,1]$, while resolving convergence issues, is to multiply $D$ by a single trainable scalar, as in~\cite{hoffer2018softmaxfix}.
 %
 
Finally, the viewpoint loss is defined as (with $\lambda=5$): 
\begin{equation}
L_{viewpoint}(X^{ref},X^+,X^-) = L_{triplet}(X^{ref},X^+,X^+) + \lambda L_{flip}(X^{ref}).
\label{eq:L_viewpoint}
\end{equation}
Table~\ref{tbl:comparison2} shows the gains yielded by different combinations of the loss functions discussed above.
The combination of the triplet loss and the flip loss (Equation~\eqref{eq:L_viewpoint}) results in the best performance.

\begin{table}[t]
\begin{center}
\begin{tabular}{|l|c|}
\hline
Loss function (real data) & Score (mAVP24)\\
\hline
 {\color{violet}Geometric loss}, Equation~\eqref{eq:geometricLoss} & 43.2 \\
 {\color{violet}Contrastive loss}, Equation~\eqref{eq:contrastive} & 42.5 \\
 {\color{violet}Flip loss}, Equation~\eqref{eq:Lflip} &  43.6 \\
 {\color{violet}Triplet + Geometric loss}, Equation~\eqref{eq:Ltriplet}+\eqref{eq:geometricLoss} & 44.1 \\
 {\color{violet}Viewpoint loss}, Equation~\eqref{eq:L_viewpoint} & 44.4\\
\hline
\end{tabular}
\end{center}
\vspace*{-0.2cm}
\caption{{\textbf{Results gained by different loss functions.}
Equation~\eqref{eq:L_viewpoint}} gives the best performance compared to a variety of loss functions.}
\label{tbl:comparison2}
\vspace*{-0.6cm}
\end{table}

\vspace{0.1in}
\noindent{\bf Putting it all together:}
Finally, the whole network is trained on the sum of all the loss functions from Figure~\ref{fig:architecture}:
\begin{equation}
L_{total} = L^{RPN}_{classification} + L^{RPN}_{regression} + L^{Classifier}_{classification} + L^{Classifier}_{regression}+L_{viewpoint}.
\label{eq:LTotal}
\end{equation}
The first four terms are from~\cite{ren2015faster} and the last term is from Equation~\eqref{eq:L_viewpoint}.


\section{Results}
\label{section:results}
Our evaluation is performed on PASCAL3D+~\cite{xiang_wacv14pascal3d}, which contains 
manually-annotated images from VOC2012~\cite{everingham2010pascalVOC12} \& ImageNet~\cite{deng2009imagenet}. 
All the experiments were conducted using the Keras~\cite{chollet2015keras} framework with TensorFlow~\cite{tensorflow2015-whitepaper} backend.

\begin{figure}[t]
\setlength{\tabcolsep}{.01em}
\centering
\begin{tabular}{cccc}
\includegraphics[height=0.15\textheight]{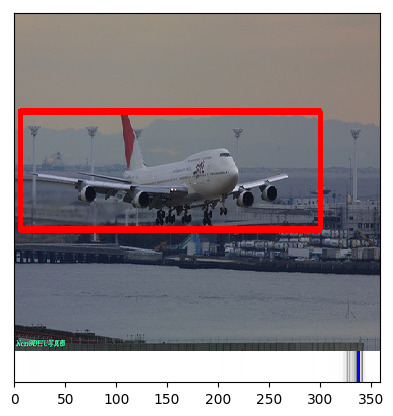}
&
\includegraphics[height=0.15\textheight]{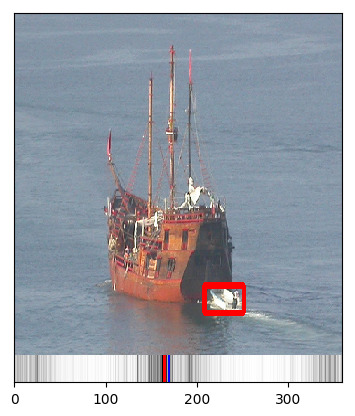}
&
\includegraphics[height=0.15\textheight]{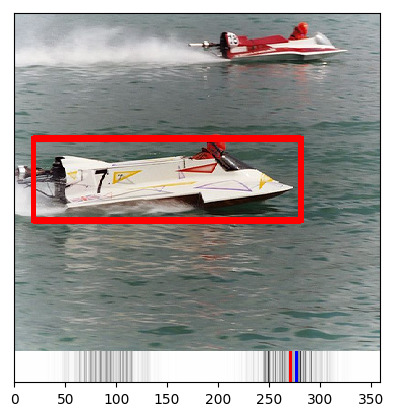}
&
\includegraphics[height=0.15\textheight]{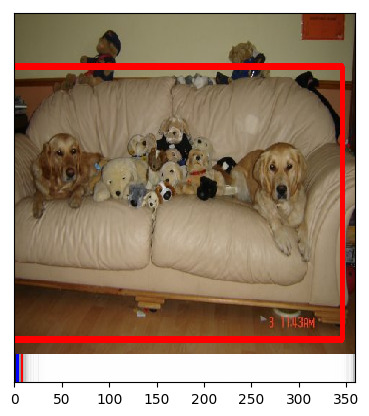}
\\
\includegraphics[height=0.15\textheight]{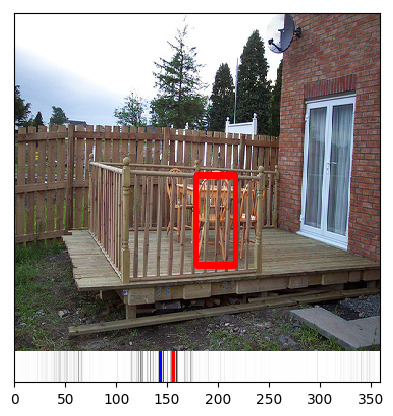}
&
\includegraphics[height=0.15\textheight]{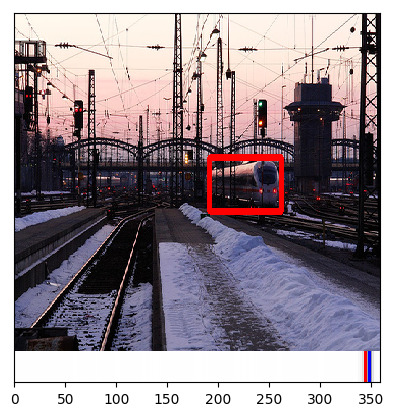}
&
\includegraphics[height=0.15\textheight]{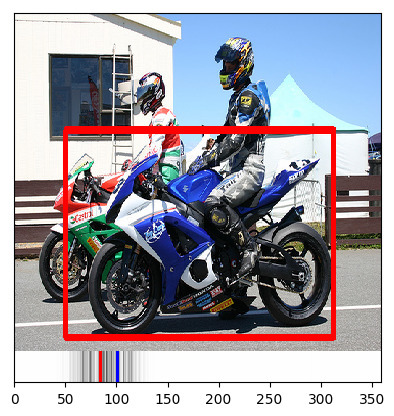}
&
\includegraphics[height=0.15\textheight]{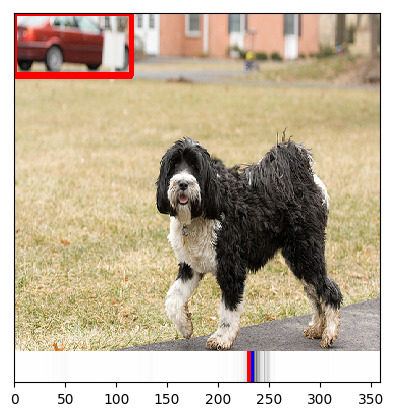}
\\
\includegraphics[height=0.15\textheight,width=0.23\textwidth]{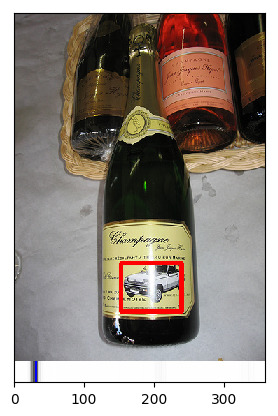}
&
\includegraphics[height=0.15\textheight,width=0.23\textwidth]{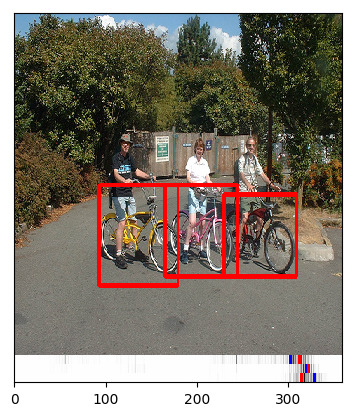}
&
\includegraphics[height=0.15\textheight,width=0.23\textwidth]{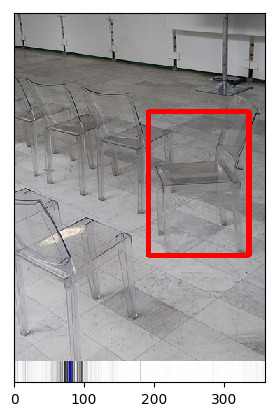}
&
\includegraphics[height=0.15\textheight,width=0.23\textwidth]{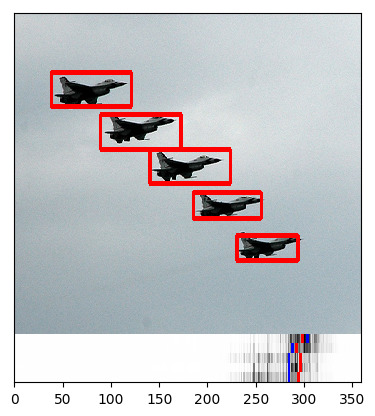}
\\
\end{tabular}
\vspace*{-0.3cm}
    \caption{\textbf{Correct viewpoint estimation predictions}. 
    The bar below each image ($0^\circ-360^\circ$) indicates in blue our highest viewpoint prediction, in red the ground truth, and in black predictions with high confidence.
Our prediction and the ground truth fall in the same bin, i.e., within $15^\circ$ from each other.}
    \label{fig:good_images}%
\end{figure}

Figure~\ref{fig:good_images} shows examples of correct predictions made by our model, as well as the detected bounding boxes.
The bar below each image indicates in blue our highest viewpoint prediction, in red the ground truth, and in black predictions with high confidence.
It can be seen that in most cases our prediction falls in the same bin as the ground truth.
Moreover, in most cases the predictions with high confidence (in black) are nicely clustered.
The exception is the image of the boats, for which two clusters of $180^\circ$-difference are evident.
This can be explained by the horizontal near-symmetry of the object.


The common evaluation metric for this problem is the {\em mean Average View Precision (mAVP)}~\cite{xiang_wacv14pascal3d}.
Briefly, in AVP, the output from the detector is considered to be correct if the bounding boxes' overlap is larger than 50\%
{\bf AND} the viewpoint is correct.
The  AVP is defined as the area under the Viewpoint Precision-Recall (VPR) curve. 
It is therefore a joint metric both for detection and for viewpoint estimation.
Following previous work, we compare our results based on the discrete AVP with $K=24$ viewpoint bins.

\subsection{Training}
\label{subsec:training}

Our model was initialized with weights from Faster R-CNN, trained on VOC2012 \& VOC2007 datasets.
The weights of the viewpoint estimation head were initialized with Xavier initialization~\cite{glorot2010Xavier}.
Adam optimizer~\cite{kingma2014adam} was used with the learning rate set to $lr=10^{-4},\beta_1 = 0.9 ,\beta_2 = 0.999$, unless otherwise specified. 

Each training step was performed using a single image, from which we took a mini-batch of 32 region proposals, out of the proposals the network had made.
By default, half of the regions included the objects and half did not;
however if the network did not provide enough regions of objects, we padded the mini-batch with more background regions.

We started the training with the synthetic data, followed by the real data.
For the synthetic data, we created $\sim$$100K$ synthetic images per category.
We fixed the weights of the detection \& classification network, since we noticed that synthetic data decreased the detection results significantly. 
We fine-tuned only the viewpoint estimation head, training for $200K$ iterations.

As real data, we used the $22K$ training images from the annotated data of PASCAL3D+.
We augmented the data by horizontally-flipped images.   
We started the training with the weights gained from the synthetic training and fine-tuned the whole network.
Our model was trained for $200K$ iterations.
Then, we reduced the learning rate by a factor of $10$ and continued the training only for the viewpoint estimation head, for $150K$ iterations.

\vspace{-0.2in}
\subsubsection{Training with our triplet loss.}
At every iteration, we randomly selected a class and a reference image from this class. 
As a positive example, an image from the same class, whose viewpoint is within $5^\circ$ from the reference, was chosen.
As a negative example, we started from "easy" images (from the same class, but faraway viewpoints) and worked our way to more difficult ones. 
Specifically, for the first 100K triplets, we sampled the distance to the reference from a Gaussian centered at $100^\circ$ with std of $20^\circ$ and selected a suitable image.
After the loss has stabilized, we sampled from a Gaussian centered at $15^\circ$ with std of $2^\circ$.

\begin{figure}[t]
\setlength{\tabcolsep}{.01em}
\centering
\begin{tabular}{cc}
\includegraphics[width=.45\textwidth]{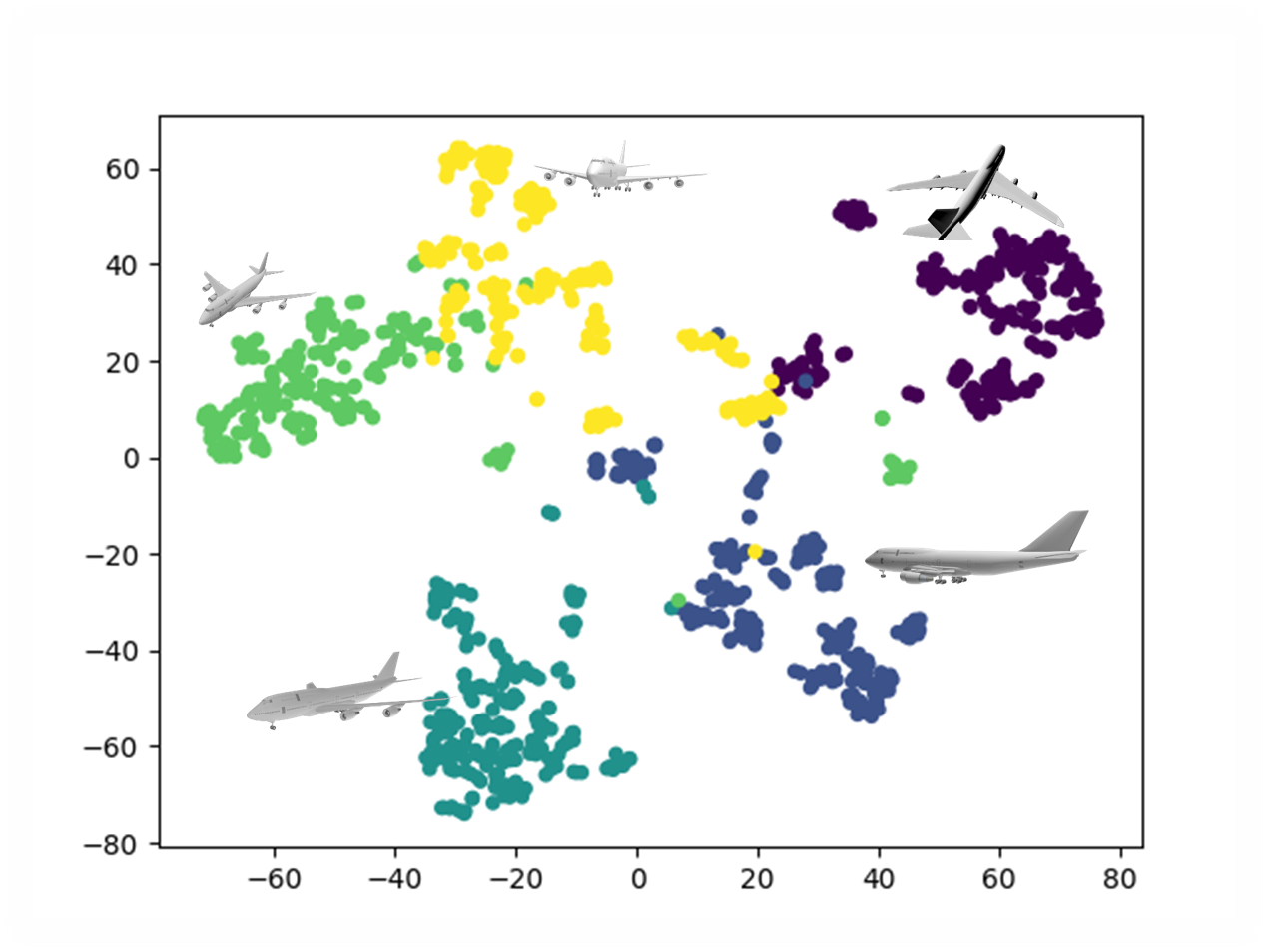}&
\includegraphics[width=.45\textwidth]{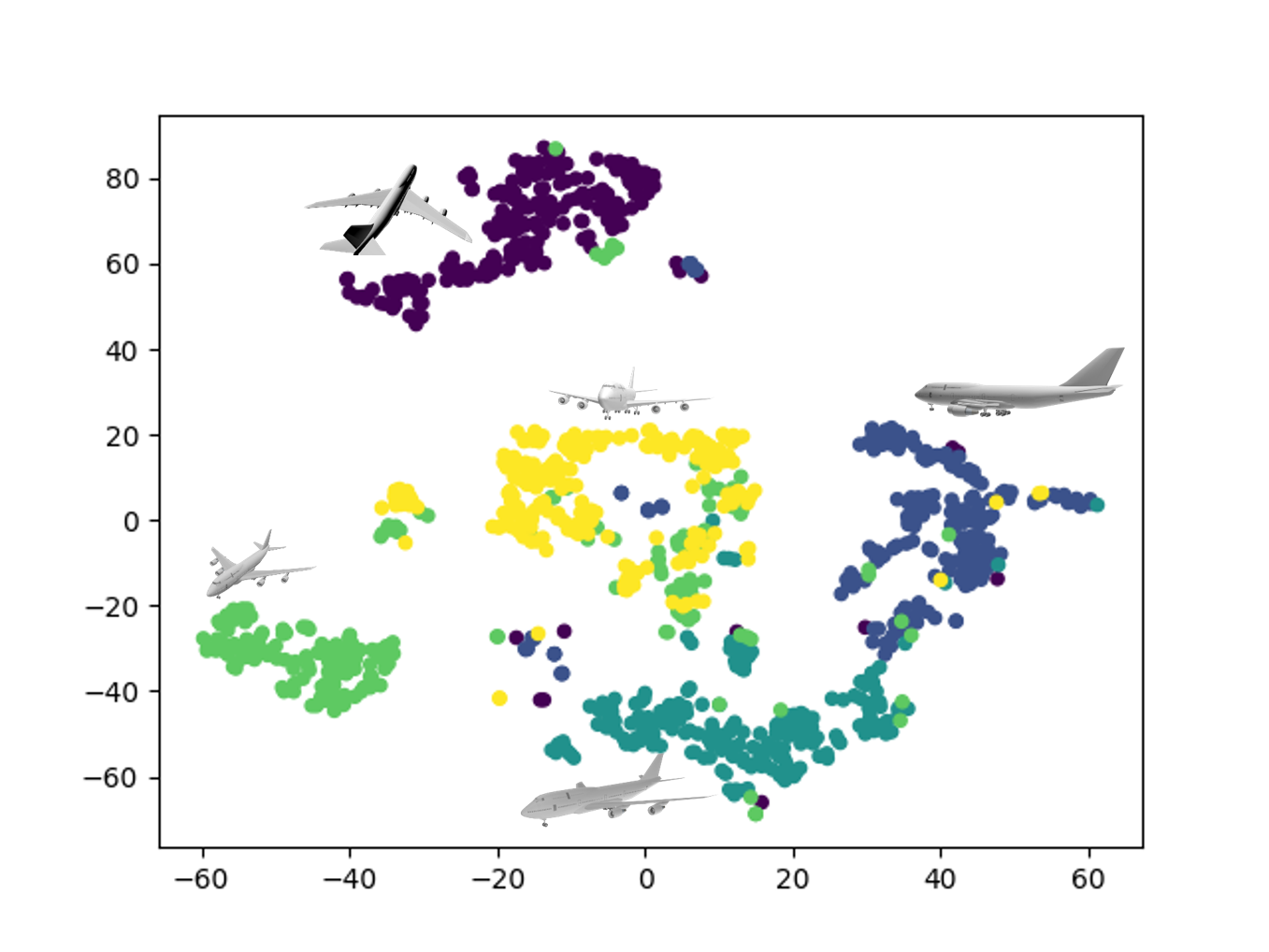}\\
(a) Triplet loss embedding & (b) Geometric loss embedding
\end{tabular}
\vspace*{-0.3cm}
    \caption{\textbf{Embeddings.} 
    Each point represents an image embedding  in the feature space (using t-SNE), where color corresponds to the ground-truth viewpoint bin. 
    The triplet loss manages not only to better separate the viewpoint bins, but also to better reflect the circular nature of the problem, than the geometric loss.}
    \label{fig:tsne}
\end{figure}

Figure~\ref{fig:tsne} shows the 2D embedding of the airplane class, using t-SNE~\cite{maaten2008tsne}, as resulted from the triplet network.
Similar viewpoints are better clustered when using our loss than when using the  geometric loss of~\cite{su2015renderforCNN}. 
Moreover, the points are structured in a more "circular" shape,  which reflects the circular nature of our problem, as explained in~\cite{su2015renderforCNN}.
Thus, the triplet loss not only better separates the embeddings, but also manages to push the features outward.   

\vspace{-0.2in}
\subsubsection{Training with videos.}
The key idea behind the use of videos is that though the viewpoint is unknown, we do know that viewpoints of subsequent frames should be similar.
To realize this idea, we use the triplet architecture, this time within a semi-supervised learning scheme.
We downloaded $100$ unannotated YouTube videos that contain objects from our categories, for which it is  unknown whether the object appears in a frame or not.
All these videos have large motion, such as a landing airplane or bike racing.
In addition, $10$ videos per class were utilized, each
 containing a single object that rotates by $360^\circ$.
 For them, it is guaranteed that the object appears in all the frames and that  all the viewpoints are sampled.
Each video is a  few minutes long, containing thousands of frames.

At each training iteration, a triplet was chosen, where the reference frame was randomly selected from some video, the positive frame was its adjacent frame (assuming that the viewpoint did not change much) and the negative frame was taken from a later frame in this video.
The weights were initialized to the results of the  regular triplet loss discussed above.
The only labeling performed was the class of the object in the video and an estimation of the gap needed for the negative frame.
We note that our viewpoint loss function is a combination of the flip loss and the triplet loss (Equation~\eqref{eq:L_viewpoint}), yet videos are not associated with labels that allow us to compute the flip loss.
Therefore, when using videos, the flip loss term used a random real image, rather than a video frame. 

\subsection{Comparison with state-of-the-art results}
\label{subsection:comparison}
Table~\ref{tbl:comparisonAll} "zooms into" the findings of Table~\ref{tbl:comparison_summary}, showing the gains for the different classes, attributed to the different components of our model.
The upper part shows the results of previous work.
The middle part shows the results of applying Insights 1,4,5.
In particular, when replacing VGG by ResNet and maintaining the same loss and data as~\cite{massa2016crafting}, the results improve from $36.1$ of~\cite{massa2016crafting} to $39.5$, using both real and synthetic data (and to $37.6$ using only real data). 
By using the geometric loss from Equation~\eqref{eq:geometricLoss}, the performance is improved to $40.6$.
When choosing the bin that integrates on the distribution of the viewpoints within it (Equation~\eqref{eq:viewpoint_score}), instead of choosing the maximum activation bin, the result is further improved to $43.2$.
This nice improvement can be explained by noting that the bin integral method can be considered as noise reduction, which is beneficial especially for noisy classes, such as bicycles and motorbikes.

The lower part of the table shows the influence of Insights 2,3.
From now on we assume that our model uses {\color{teal}ResNet/Faster R-CNN}, was trained on {\color{olive}synthetic \& real data} and chooses the bin using an {\color{RedOrange} integral} on the distribution.
Our viewpoint loss improves the results by $1.2\%$.
The video data further improves the result  by $1.5\%$.
Overall we achieved improvement of $9.8\%$ compared to the current state-of-the-art results.

\begin{table}[t]
\begin{center}
\resizebox{\textwidth}{!}{\begin{tabular}{|l|c|c|c|c|c|c|c|c|c|c|c|c|}
\hline
\hspace{1.1in}Method & aero & bicycle & boat & bus & car & chair & table& mbike &sofa & train & TV & mAVP24\\
\hline\hline
{\color{RoyalPurple}\cite{su2015renderforCNN}}: {\color{teal}AlexNet/R-CNN}-{\color{violet}Geometry}-{\color{olive}synthetic+real} & 21.5 & 22.0 & 4.1 & 38.6 & 25.5 & 7.4 & 11.0 & 24.4 & 15.0 & 28.0 & 19.8 & 19.8 \\
{\color{RoyalPurple}\cite{tulsiani2015viewpoints}}: {\color{teal}VGG/R-CNN}-{\color{violet}Classification}-{\color{olive}real} & 37.0 & 33.4 & 10.0 & 54.1 & 40.0 & 17.5 & 19.9 & 34.3 & 28.9 & 43.9 & 22.7 & 31.1 \\
{\color{RoyalPurple}\cite{massa2016crafting}}: {\color{teal}VGG/Fast R-CNN}-{\color{violet}Classification}-{\color{olive}synthetic+real}& 43.2 & 39.4 & 16.8 & 61.0 & 44.2 & 13.5 & 29.4 & 37.5 & 33.5 & 46.6 & 32.5 & 36.1 \\
\hline
\hline
{\color{RoyalPurple}Ours}: {\color{teal}ResNet/Faster}-{\color{violet}Classification}-{\color{olive}real} & 41.6&33.7&20.6&65.3 & 45.4& 17.9& 33.8& 36 & 34.5 & 48.6&   36.6&37.6  \\
{\color{RoyalPurple}Ours}: {\color{teal}ResNet/Faster}-{\color{violet}Classification}-{\color{olive}synthetic+real} &43.6 & 37.1  & 19.9  & 68.5  & 48.6  & 19.8  & 37.1  & 34.2  &  38.2 & 48.3  & 39.6  & 39.5 \\
{\color{RoyalPurple}Ours}: {\color{teal}ResNet/Faster}-{\color{violet}Geometry}-{\color{olive}synthetic+real} & 43.9& 35.4 & 20.9 & 70.3 & 51.5 & 20.0 & \ 38.6 & 34.0 & 41.6 & 50.4 &40.0  & 40.6 \\
{\color{RoyalPurple}Ours}: {\color{teal}ResNet/Faster}-{\color{violet}Geometry}-{\color{olive}synthetic+real}-{\color{RedOrange}integral} & 43.5& 41.2 & 23.9 & 68.4 & 52.7 & 22.4 &41.9 & 42.0 & 44.1 &50.3 & 45.0 & 43.2 \\
\hline
\hline
{\color{RoyalPurple}Ours}: all the above +{\color{violet}Viewpoint loss} &46.6& 41.1 & \textbf{23.9} & 72.6 & 53.5 & 22.5 & 42.6 & 42.0 & 44.2 &\textbf{54.6} &44.8 & 44.4 \\
{\color{RoyalPurple}Ours}: all the above +{\color{violet}Viewpoint loss}-{\color{olive}video data} &\textbf{47.7} &\textbf{42.5}&23.8&\textbf{74.8}& \textbf{54.7}& \textbf{25.9}&\textbf{42.8}&\textbf{43.5}&\textbf{46.3}&54.6&\textbf{47.9}&\textbf{45.9}\\
\hline

\end{tabular}}
\end{center}
\caption{{\bf Our results greatly improve SOTA results.} 
The total gain is improved by $9.8\%$ on PASCAL3D+, from 36.1 of \cite{massa2016crafting} to 45.9.
}
\label{tbl:comparisonAll}
\vspace*{-0.3cm}
\end{table}

It is interesting to note that different methods improve different categories.
For instance, the integration method vastly improves the motorbike and the bicycle classes.
In these classes there are many images that contain more than one object from the class, which are very close to one another.
Their detected bounding boxes overlap and objects produce "noisy viewpoints" for others.
When integrating over all the bounding boxes, some of which do contain a single object, this noise is reduced.
Moreover, the viewpoint distribution in PASCAL3D+ for these classes is more uniform than in most other classes.
Hence, our network has no bias and tends to assign probabilities to all the object's symmetries.
This is beneficial for the bin integral method, since collecting more information improves viewpoint prediction.

The flip loss shows improvements mainly for the "rectangular" objects, such as buses, trains and tables.
We infer that flipping indeed helps the network in resolving some of the symmetry ambiguities, as desired.  

The viewpoint (triplet/flip) loss is mostly beneficial for classes for which the geometric loss errs by $180^\circ$ (object facing backward/forward), such as airplanes, buses and trains.
A possible explanation is that unlike the geometric loss, which relates between close viewpoints through its use of  Gaussian weights, the viewpoint loss relates also between faraway viewpoints.

The use of video clips improves 
 the classes for which we had videos that contain almost all viewpoints.

\vspace{-0.2in}
\subsubsection{Limitations:}
\begin{figure}[t]
\setlength{\tabcolsep}{.01em}
\centering
\begin{tabular}{cccc}
\includegraphics[width=0.2\textwidth]{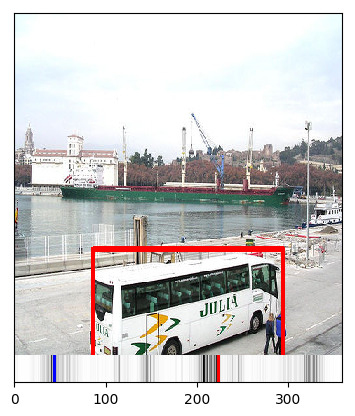}
&
\includegraphics[width=0.2\textwidth]{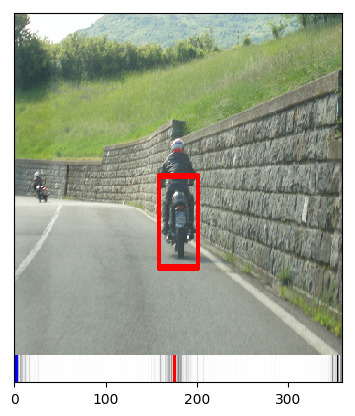}&
\includegraphics[width=0.2\textwidth]{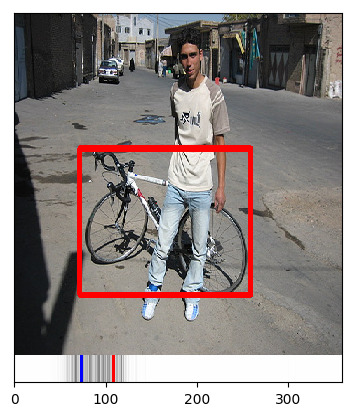}&
\includegraphics[width=0.2\textwidth]{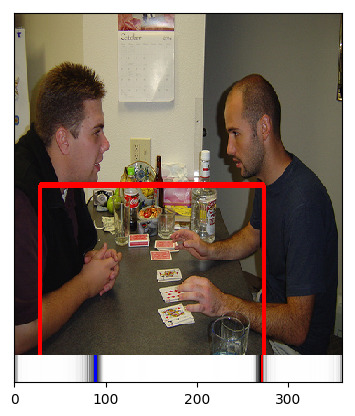}\\
\end{tabular}
\vspace*{-0.3cm}
    \caption{\textbf{Typical false viewpoint estimation cases.} 
    Our prediction for the bus and the motorbike is $180^\circ$-opposite; for the bicycle  our prediction matches the viewpoint of the handlebar rather than that of the main frame;
    the table should have two correct viewpoints, of which our viewpoint estimator chose one.
}
\vspace*{-0.3cm}
    \label{fig:bad_images}%
\end{figure}

Figure~\ref{fig:bad_images} illustrates some typical failures.
For the bus and the motorbike, the failures are due to backward/forward symmetry---our model predicted the $180^\circ$-opposite viewpoint.
The false prediction for the bicycle is due to the handlebar position, which is not aligned with the main frame.
%
The table illustrates a case where two viewpoints are equally correct (i.e. there is no front \& back to a rectangular table), yet our algorithm chose one viewpoint whereas the ground truth is the other.
\section{Conclusions}
\label{section:conclusions}

This paper has addressed the problem of viewpoint estimation of an object in a given image. 
It provides five main insights, which regard all the components of the network: architecture, training data, the loss function, and the integration of the results.

Based on these insights, a network was designed such that:
(i) The architecture jointly solves detection, classification, and pose estimation, using the most advanced CNN for performing the two former tasks.
(ii) To handle the shortage in labeled data, the paper proposes to add both videos and flipped images to the training stage.
(iii) A novel loss function that takes into account both the geometric nature of the problem, as well as the constraints posed by videos and flipped images, is introduced.
(iv) While previous works predicted the viewpoint using the maximum activation, we propose an integration scheme for prediction. 

Our network improves the state-of-the-art results for this problem  on PASCAL3D+ by 9.8\%.
The paper carefully analyzes the influence of each component on the overall performance.
The code will be  released upon publication.


\vspace{0.1in}
\noindent
{\bf Future directions:}
Our viewpoint estimation is based only on the information within the bounding box.
However, information from the full image can be helpful.
For instance, the wave direction may assist in determining the boat's viewpoint, or passengers on the platform may indicate the train's direction.

Second, the available dataset should be enhanced.
As the improved performance due to our additional data may imply, larger datasets are likely to benefit viewpoint estimation. 
Moreover, better annotation methods are necessary, as currently some images are falsely annotated, which biases both training and testing.
Finally, for certain types of objects (e.g. circular tables or the table in Figure~\ref{fig:bad_images}), any attempt to define a single ground-truth viewpoint is doomed to fail.
Such special cases should be given proper attention.

\clearpage
{\small
\bibliographystyle{splncs}
\bibliography{egbib}
}
\end{document}